\documentclass{article}
\pdfoutput=1
\usepackage[nohyperref]{naaclhlt2019}
\usepackage{times}
\usepackage{latexsym}
\usepackage{amssymb}
\usepackage{algorithm}
\usepackage[noend]{algpseudocode}
\usepackage{amsmath}
\usepackage{amssymb}
\usepackage{graphicx}

\aclfinalcopy % Uncomment this line for the final submission

\title{Large Scale Question Paraphrase Retrieval with\\ Smoothed Deep Metric Learning} 

\author{Daniele Bonadiman \\
  University of Trento \\
  Trento, Italy \\
  {\tt d.bonadiman@unitn.it} \\\And
  Anjishnu Kumar \\
  Amazon Alexa \\
  Seattle, USA \\
  {\tt anjikum@amazon.com} \\
  \\\And
  Arpit Mittal \\
  Amazon Alexa \\
  Cambridge, UK \\
  {\tt mitarpit@amazon.com} \\}
\date{}

\begin{document}
\maketitle
\begin{abstract}

The goal of a Question Paraphrase Retrieval (QPR) system is to retrieve equivalent questions that result in the same answer as the original question. Such a system can be used to understand and answer rare and noisy reformulations of common questions by mapping them to a set of canonical forms. This has large-scale applications for community Question Answering (cQA) and open-domain spoken language question answering systems. In this paper we describe a new QPR system implemented as a Neural Information Retrieval (NIR) system consisting of a neural network sentence encoder and an approximate k-Nearest Neighbour index for efficient vector retrieval.
We also describe our mechanism to generate an annotated dataset for question paraphrase retrieval experiments automatically from question-answer logs via distant supervision. 
We show that the standard loss function in NIR, triplet loss, does not perform well with noisy labels. We propose smoothed deep metric loss (SDML) and with our experiments on two QPR datasets we show that it significantly outperforms triplet loss in the noisy label setting.% Furthermore the simplicity of its implementation makes it a strong candidate for consideration for other NIR tasks.
\end{abstract}

\section{Introduction}

In this paper we propose a Question Paraphrase Retrieval (QPR) \cite{bernhard2008answering} system that can operate at industrial scale. A QPR system retrieves a set of paraphrase questions for a given input, enabling existing question answering systems to answer rare formulations present in incoming questions. QPR finds natural applications in question answering systems, and is especially relevant to the community Question Answering (cQA) systems.

Common cQA websites such as Quora or Yahoo Answers are platforms in which users interact by asking and answering questions. The community-driven nature of these platforms leads to problems such as question duplication. Therefore, having a way to identify paraphrases can reduce clutter and improve the user experience. Question duplication can be prevented by presenting users a set of candidate paraphrase questions by retrieving them from the set of questions already in the system.

Open-domain QA systems provide answers to a user's questions with or without human intervention. These systems are employed by virtual assistants such as Alexa, Siri, Cortana and Google Assistant. Some virtual assistants use noisy channels, such as speech, to interact with users. Questions that are the output of an Automated Speech Recognition (ASR) system could contain errors such as truncations and misinterpretations. Transcription errors are more likely to occur for rarer or grammatically non-standard formulations of a question. For instance `Where Michael Jordan at?' could be a reformulation for `Where is Michael Jordan?'. A QPR system tries to mitigate the impact of this noise by identifying an answerable paraphrase of the noisy query and hence improves the overall performance of the system. 

%Despite similarities, QPR task differs from the better known Paraphrase Identification (PI) task. % for several reasons.
Paraphrase Identification (PI) \cite{mihalcea2006corpus,islam2009semantic,he2015multi} is a related task where the objective is to recognize whether a pair of sentences are paraphrases. The largest dataset for this task was released by Quora.com\footnote{https://data.quora.com/First-Quora-Dataset-Release-Question-Pairs}. State-of-the-art approaches on this dataset use neural architectures with attention mechanisms across both the query and candidate questions \cite{parikh2016decomposable,wang2017bilateral}. %This results in high performance, 
%However these systems cannot be applied for paraphrase retrieval since that would involve comparison of every pair of questions. anjikum: Our system compares every pair as well, only that the comparison is very efficient because each question is a vector.
%Since it involves vector comparison for every question pair, these systems are impractical for large scale paraphrase retrieval applications.anjikum : We are still 
However these systems are impractical for large scale applications with millions of candidates, since they involve quadratic number of vector comparisons per question pair, which are non-trivial to parallelize.

Information Retrieval (IR) systems have been very successful to operate at scale for such tasks. However, standard IR systems, such as BM25 \cite{bm25}, are based on lexical overlap rather than on deep semantic understanding of the questions \cite{robertson2009probabilistic}, making them unable to recognize paraphrases that lack significant lexical overlap. In recent years, the focus of the IR community has moved towards neural network based systems that can provide a better representation of the object to be retrieved, while maintaining the performance of the standard model. Neural representations can capture latent syntactic and semantic information from the text, overcoming the shortcomings of systems based purely on lexical information.  Moreover, representations trained using a neural network can be task specific, allowing them to encode domain specific information that helps them outperform generic systems.

The major components of a Neural Information Retrieval (NIR) system are a neural encoder and a k-Nearest Neighbour (kNN) index \cite{mitra2017neural}. The encoder is a neural network capable of transforming an input example, in our case a question, to a fixed size vector representation. In a standard setting, the encoder is trained via triplet loss \cite{schroff2015facenet} to reduce the distance between two paraphrase vectors as compared to a paraphrase vector and a non-paraphrase vector. After being trained for this task, the encoder is used to embed the questions that can be later retrieved at inference time. The encoded questions are added to the kNN index for efficient retrieval. The input question is encoded and used as a query to the index, returning the top k most similar questions

Public datasets, such as Quora Question Pairs, are built to train and evaluate classifiers to identify paraphrases rather than evaluating retrieval systems. Additionally, the Quora dataset is not manually curated, thus resulting in a dataset that contains false negative question paraphrases. This problem introduces noise in the training procedure when minimizing the triplet loss. This noise is further exacerbated in approaches for training procedures that exploit the concept of hard negatives, i.e., mining the non-paraphrase samples that are close to paraphrase samples in the vector space \cite{manmatha2017sampling}. 

In this work, we propose a loss function that minimizes the effect of false negatives in the training data. The proposed loss function %trains the model to identify a good paraphrase in a set of randomly sampled questions and 
uses label smoothing to assign some probability mass to negative examples, thus mitigating the impact of false negatives.% by forcing the network to model interactions between random sets of question pairs. 

The proposed technique is evaluated on two datasets: a distantly supervised dataset of questions collected from a popular virtual assistant system, and a modified version of the Quora dataset that allows models to be evaluated in a retrieval setting. The effect of our proposed loss and the impact of the smoothing parameters are analysed in Section 4. 

% The paper describes a paraphrasing system:
% \begin{itemize}
%     \item It finds related questions that have been answered.
% \end{itemize}

% Use cases

% \begin{itemize}
%     \item Question Answering systems with noisy input.
%     \item Community question answering websites (Quora, Yahoo Answer, ..)
% \end{itemize}

% How to paraphrase 

% \begin{itemize}
%     \item Generative Models
%     \item Retrieval Models
% \end{itemize}

% Why retrieval?

% \begin{itemize}
%     \item Generative models (query rewriting) gives no guaranties on the quality of the produced utterance
%     \item There is not an explicit way to ensure that the produced answer is answerable.
% \end{itemize}

% What is in the state of art? 

% \begin{itemize}
%     \item paraphrase identification
%     \item attention is not an option
%     \item siamese networks (Convolutional -> to cop with noisy inputs)
%     \item Approximate KNN (Exact is not feasible for big datasets) 
% \end{itemize}

% How siamese models are trained? Triplet loss

% \begin{itemize}
%     \item there are various arguments in the literature on how to choose negative examples for triplet loss
%     \item Optimal results with triplet loss are usually obtained with hard example mining 
%     \item It is not practical to retrieve hard example at every epoch when dealing with big dataset >1M utterances.
%     \item 
% \end{itemize}

% What we propose?

% \begin{itemize}
%     \item The loss.
%     \item the retrieval based framework for Quora.
% \end{itemize}

\section{Question Paraphrase Retrieval} %probably we need to change the title of the section

In QPR the task is to retrieve a set of candidate paraphrases for a given query. Formally, given a new query $q_{new}$, the task is to retrieve k-questions, $Q_k$ ($|Q_k| = k$), that are more likely to be paraphrases of the original question. The questions need to be retrieved from a given set of questions $Q_{all}$ such that $ Q_k \subseteq Q_{all}$, e.g., questions already answered in a cQA website.

\subsection{System overview}
The QPR system described in this paper is made of two core components: an encoder and an index.
The encoder $\phi$ is a function ($\phi: Q \rightarrow \mathbb{R}^{n}$) that takes as input a question $q \in Q$ and maps it to a $n$-dimensional vector representation.
The index is defined as the encoded set of all the questions that can be retrieved $\{ \phi(q') | q' \in Q_{all} \}$ using the standard kNN search mechanism.

\subsubsection{Encoder}
The encoder $\phi$ used by our system is a neural network that transforms the input question to a fixed size vector representation. To this end, we use a convolutional encoder since it scales better (is easily parallelizable) compared to a recurrent neural network encoder while maintaining similar performance on sentence matching tasks \cite{yin2017comparative}. The encoder uses a three-step process: %(i) sentence tokens are mapped to a fixed size word embeddings, (ii) a convolutional layer combines it with (local) word-ordering information, and finally (iii) a global max pooling layer is used reduce the sentence representation to a fixed size vector. These three operations are decribed below:

 (i) An embedding layer maps each word $w_i$ in the question $q$ to its corresponding word embedding $x_i \in \mathbb{R}^{e_{dim}}$ and thereby generating a sentence matrix $X_q \in \mathbb{R}^{l \times e_{dim}}$, where $l$ is number of words in the question. We also use the hashing trick of \cite{weinberger2009feature} to map rare words to $m$ bins via random projection to reduce the number of false matches at the retrieval time.
 
(ii) A convolutional layer \cite{kim2014convolutional} takes the question embedding matrix $X_q$ as input and applies a trained convolutional filter  $W \in \mathbb{R}^{e_{dim}win}$ iteratively by taking at each timestep $i$ a set of $win$ word embeddings. This results in the output:
$h^{win}_i = \sigma(W x_{i-\frac{win}{2}:i+\frac{win}{2}} + b)$, where $\sigma$ is a non linearity function, $tanh$ in our case, and $b \in \mathbb{R}$ is the bias parameter. By iterating over the whole sentence it produces a feature map $\textbf{h}^{win} = [h^{win}_1, .., h^{win}_l]$. 

(iii) A global max pooling operation is applied over the feature map ($\hat{h}^{win} = max(\textbf{h}^{win})$) to reduce it into a single feature value. The convolutional step described above is applied multiple times ($c_{dim}$ times) with varying window size with resultant $\hat{h}$ values concatenated to get a feature vector $h \in \mathbb{R}^{c_{dim}}$ which is then linearly projected to an $n$-dimensional output vector using a learned weight matrix $W_p \in \mathbb{R}^{n \times c_{dim}}$.

\subsubsection{kNN Index}
For our system we use FAISS \cite{JDH17} as an approximate kNN index for performance reasons. All the questions ($Q_{all}$) are encoded offline using the encoder $\phi$ and added to the index. At retrieval time a new question is encoded and used as a query to the index. FAISS uses a predefined distance function (e.g. Euclidean distance) to retrieve the nearest questions in the vector space.

\section{Training}
Typical approaches for training the encoder use triplet loss \cite{schroff2015facenet}. This loss attempts to minimize the distance between positive examples while maximizing the distance between positive and negative examples.

The loss is formalized as follows:
\begin{equation}
    \sum_i^N[ \lVert \phi(q_i^a) - \phi(q_i^p) \rVert^2_2 - \lVert \phi(q_i^a) - \phi(q_i^n) \rVert^2_2 + \alpha]_+
\end{equation}

where $q_i^a$ is a positive (anchor) question, $q_i^p$ is a positive match to the anchor (a valid paraphrase), $q_i^n$ is a negative match (i.e. a non-paraphrase), $\alpha$ is a margin parameter and $N$ is the batch size. 

In a recent work by \citealt{manmatha2017sampling}  the authors found that better results could be obtained by training the above objective with hard negative samples. These hard negatives are samples from the negative class that are the closest in vector space to the positive samples, hence most likely to be misclassified.

However, in our case, and in other cases with noisy training data, this technique negatively impacts the performance of the model since it starts focusing disproportionately on any false negative samples in the data (i.e. positive examples labelled as negative due to noise) making the learning process faulty.

\subsection{Smoothed Deep Metric Learning}
In this paper we propose a new loss function that overcomes the limitation of triplet loss in the noisy setting. Instead of minimizing the distance between positive examples with respect to negative examples, we view the problem as a classification problem. Ideally we would like to classify the paraphrases of the original question amongst all other questions in the dataset. This is infeasible due to time and memory constraints. We can however approximate this general loss by identifying a valid paraphrase in a set of randomly sampled questions \cite{kannan2016smart}.
We map vector distances into probabilities similar to \citealt{goldberger2005neighbourhood} by applying a softmax operation over the negative squared euclidean distance:

\begin{equation}
    \hat{p}{(a,i)} = \frac{e^{-\lVert \phi(q^a) - \phi(q^i) \rVert^2_2}}{\sum_j^{N} e^{-\lVert \phi(q^a) - \phi(q^j) \rVert^2_2}}
\end{equation}

where $q^a$ is an anchor question and $q^j$ and $q^i$ are questions belonging in a batch of size $N$ containing one paraphrase and $N-1$ randomly sampled non-paraphrases.The network is then trained to assign a higher probability to pair of questions that are paraphrases. 

Additionally, we apply the label smoothing regularization technique \cite{szegedy2016rethinking} to reduce impact of false negatives. This technique reduces the probability of the ground truth by a smoothing factor $\epsilon$ and redistributes it uniformly across all other values, i.e.,
 
 \begin{equation}
 p'(k|a) = (1 - \epsilon ) p(k|a) + \frac{\epsilon}{N}
 \end{equation}
 where $p(k|a)$ is the probability for the  gold label.
 The new smoothed labels computed in this way are used to train the network using Cross-Entropy (CE) or Kullback - Leibler (KL) divergence loss.\footnote{In this setting CE loss and KL divergence loss are equivalent in expected values. However, we use the KL divergence loss for performance reasons.} A standard cross-entropy loss tries to enforce the euclidean distance between all random points to become infinity, which may not be feasible and could lead to noisy training. Instead, assigning a constant probability to random interactions tries to position random points onto the surface of a hypersphere around the anchor.
 
\begin{figure}
\centering
  \includegraphics{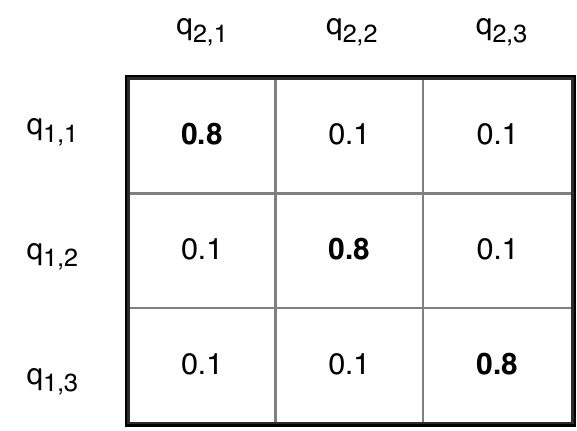}
  \caption{Batched implementation of the loss with smoothing parameter $\epsilon = 0.3$ and batch size $N = 3$. Each paraphrase pair $<q_{1,j}, q_{2,j}>$ in the batch is compared with all the others questions in the batch. }
  \label{fig:loss}
\end{figure}

The sampling required for this formulation can be easily implemented in frameworks like PyTorch \cite{paszke2017automatic} or MxNet \cite{chen2015mxnet} using a batch of positive pairs $<q_{1,j}, q_{2,j}>$ derived from a shuffled dataset, as depicted in Figure~\ref{fig:loss}. In this setting, each question $q_{1,i}$ would have exactly one paraphrase, i.e., $q_{2,i}$ and $N-1$ all other questions $q_{2, j}$ when $j \neq i$ would serve as counter-examples. This batched implementation reduces training time and makes sampling tractable by avoiding sampling $N$ questions for each example, reducing the number of forward passes required to encode the questions in a batch from $\mathcal{O}(N^2)$ in a naive implementation to $\mathcal{O}(2N)$.  %by operating on shuffled data in a mini-batch of size $N$, wherein the $N-1$ other random paraphrase pairs in the mini-batch can be treated as negative samples for the purpose of computing this loss.
% The authors argue that the proposed approach can help to reduce overfitting by not assigning the full probability to the ground truth label and it can avoid the model to become too confident about its predictions. The second point is particularly relevant for our setting.

\section{Experiments}

In this section, we present the experimental setup used to validate our approach for QPR using the Smoothed Deep Metric Learning (SDML) technique. %Section~\ref{datasets} describes the datasets used for evaluation, that are the Quora dataset for community question answering and the Open Domain QA dataset that was built by mining logs from an underlying high precision Question Answering system.

\subsection{Datasets}

In order to generate a dataset for question paraphrase retrieval we propose a technique that uses distant supervision to create it automatically from high-precision question-answer (QA) logs. Due to the proprietary nature of our internal dataset, we also report numbers on a modified version of Quora paraphrase identification dataset that has been adapted for the paraphrase retrieval task.

\label{datasets}
\subsubsection{Open Domain QA dataset}

Our open domain Q\&A dataset is created by weak supervision method using high precision QA logs of a large scale industrial virtual assistant. % such as Alexa, the Google Assistant, Siri or Cortana. 
%We constrain ourselves to deterministic, high precision question answering subsystems so that we can assume that there is no noise in the mapping from a question to an answer.
From the logs we retrieve `clusters' of questions that are mapped to the same answer. However we notice that this may generate clusters where unrelated questions are mapped to a generic answer. For instance many different math questions may map to the same answer; a given number. To further refine these clusters, the data is filtered using a heuristic based on an intra-cluster similarity metric that we call cluster \textit{coherence}, denoted as $c$. We define this metric as the mean Jaccard similarity \cite{levandowsky1971distance} of each question in a cluster to the cluster taken as the whole. 

Mathematically, for a given cluster $ \mathbb{A} =\{q_1, q_2 ... q_n\}$ 
and defining $\mathbb{T}_{q_i} = \{w_{i_1}, w_{i_2}, ... w_{i_k}\}$ as shorthand for the set of unique tokens present in a given question, the coherence of the cluster is defined as: 

\begin{equation}
    \mathbb{S} = \bigcup_{i=1}^{n} \mathbb{T}_{q_i}
\end{equation}
\begin{equation}
    c = \frac{1}{n}\Sigma_{i=1}^{n} \frac{|\mathbb{T}_{q_i}\cap \mathbb{S}|} {|\mathbb{S}|}
\end{equation}

In practice we found that even a small coherence filter ($c < 0.1$) is able to eliminate all incoherent question clusters. % - which generally contain unrelated questions mapping to a generic answer. For instance many different math questions may map to the same answer; a given number. 
Our approach to weak supervision can be considered as a generalized instance of the candidate-generation noise-removal pipeline paradigm used by \citealt{kim2018efficient}. Once the incoherent clusters are removed from the dataset, the remaining clusters are randomly split in a 80:10:10 ratio into training, validation and test sets and question pairs are generated from them\footnote{The open domain QA dataset contains on order of 100k - 1M training clusters, 10-100k clusters each for validation and testing, and a search index of size $\approx 10M$.}. A second filter is applied to remove questions in the validation and test sets that overlap with questions in the training set. The final output of the weak supervision process is a set of silver labelled  clusters with $>99\%$ accuracy based on spot checking a random sample of 200 clusters. % that all questions in the cluster are valid paraphrases of each other. 
%However, sometimes questions with similar semantic meaning may fall into different clusters due to minor changes in how the answer is worded, or because the answer to a specific question changes across temporal or geographical dimensions. Despite these limitations, metrics computed on this distantly supervised dataset correlate well with performance on the live system.

\subsubsection{Quora dataset}

We introduce a variant of the Quora dataset for QPR task. The original dataset consists of pairs of questions with a positive label if they are paraphrases, and a negative label if they are not. We identify question clusters in the dataset by exploiting the transitive property of the paraphrase relation in the original pairs, i.e., if $q_1$ and $q_2$ are paraphrases, and $q_2$ and $q_3$ are paraphrases then $q_1$ and $q_3$ are also paraphrases, hence $q_1$, $q_2$, and $q_3$ belong to the same cluster. After iterating over the entire dataset we identified $60,312$ question clusters. The question clusters are split into the training, validation and test sets such that the resulting validation and test set contains roughly $5,000$ question pairs each, and the training set contains $219,369$ question pairs\footnote{The code to generate the splits will be released upon acceptance.}. The kNN index is composed of all the questions in the original Quora datasets (including questions that appear only as negative, thus not being part of any cluster) for a total of $556,107$ questions. %During end-to-end testing of the system,i.e., the encoder and the index, we verify whether the retrieved questions belong to the same cluster.

\subsection{Experimental setup}

We described the architecture of our encoder previously in section 2.1.1. For experimentation we randomly initialized word embeddings. The size of vocabulary for Quora dataset is fixed at 50,000 whereas for the open domain QA dataset we used a vocabulary of size 100,000. To map rare words we use 5,000 bins for the Quora dataset and 10,000 bins for the QA dataset.

We set the dimensionality of word embeddings at 300 (i.e., $e_{dim} = 300$); the convolutional layer uses a window size of $5$ (i.e., $win = 5$) and the encoder outputs a vector of size $n = 300$. For triplet loss the network is trained with margin $\alpha = 0.5$. %For SDML the default number of candidate paraphrases is 512 (i.e., $N=512$)
The default batch size for all the experiments is 512 (i.e., $N=512$) and the smoothing factor for SDML, $\epsilon$, is 0.3. For all experiments training is performed using the Adam optimizer with learning rate $\lambda = 0.001$ until the  model stops improving on the validation test, using early stopping \cite{prechelt1998early} on the ROC AUC metric \cite{bradley1997use}. 

\paragraph{Evaluation.} We use \textit{IVF2000, Flat} configuration of the FAISS library as our index, which is a hierarchical index consisting of an index of k-means centroids as the top-level index. For evaluation we retrieve $20$ questions %with $10$ probes into the index each returning a pair of paraphrase questions, 
with an average query time of $<10$ ms. These questions are used to measure the system performance via standard information retrieval metrics, Precision@N ($P@N$) and Mean Reciprocal Rank (MRR). $P@N$ measures if at least one question in the first $N$ that are retrieved is a paraphrase and MRR is the average reciprocal rank (position) at which the first retrieved paraphrase is encountered.

\subsection{Results}

In the first set of experiments we measured the impact of varying the smoothing factor $\epsilon$. The results for the Quora validation set are presented in Table~\ref{epsilonsearch}. %the biggest impact results shows a difference of $\sim 4$ P@1 when including the smoothing parameter in the loss. 
We observe that the presence of smoothing leads to a significant increase over the baseline (simple cross entropy loss), and increasing this parameter has a positive impact up to $\epsilon = 0.3$.

In our second experiment, we hold the $\epsilon$ constant at $0.3$ and experiment with varying the number of negative samples. Table~\ref{Batch Size} shows the effect of an increase in the number of negative examples in a batch. The model's performance reaches its maximum value at $N=512$, i.e., with $511$ negative samples for each positive sample. We would like to point out that we limited our exploration to 1024 due to memory constraints. However, better performance may be achieved by further increasing the number of examples, since the batch becomes a better approximation of the true distribution. %It is important to notice that in this case we had to limit the number to $1024$ for memory reasons but it could be possible that the performance improves by further increasing the number of negatives, since the batch becomes a better approximation of the true distribution of questions in the index.

Table~\ref{quora_dev} and~\ref{quora_test} compare the proposed loss with the triplet loss with random sampling, TL(Rand). We compared the proposed approach with two variants of triplet loss that uses different distance functions Euclidean Distance (EUC) and Sum of Squared Distances (SSD). The euclidean distance is the standard distance function for triplet loss implementation present in popular deep learning frameworks, PyTorch and Mxnet, whereas SSD is the distance function used in the original paper of  \citealt{schroff2015facenet}. Our approach improves over the original triplet loss considerably on both datasets. The SSD distance also outperforms the EUC implementation of the loss. %implemented in popular libraries 
%on both datasets we considered.

\begin{table}[t!]
\begin{center}
\begin{tabular}{|l|rcl|}
\hline \bf $\epsilon$ & \bf P@1 & \bf P@10 & \bf MRR \\ \hline
0& 0.5568  & 0.7381 & 0.6217\\
0.1 & 0.5901 & 0.7841& 0.6591 \\
0.2 & 0.6030  & 0.8090 & 0.6762\\
0.3 & \textbf{0.6133} & 0.8113 & \textbf{0.6837} \\
0.4 & 0.6107 & \textbf{0.8144} & 0.6815 \\
\hline
\end{tabular}
\end{center}
\caption{\label{epsilonsearch} Impact of smoothing factor $\epsilon$ on the Quora validation set.}
\end{table}

\begin{table}[t!]
\begin{center}
\begin{tabular}{|l|rcl|}
\hline \bf N  & \bf P@1 & \bf P@10 & \bf MRR \\ \hline
32 & 0.5389  & 0.7444 & 0.6103\\
64  & 0.5710&0.7726& 0.6410 \\
128  & 0.6093 &0.8085 & 0.6777 \\
256  &0.6112 &\textbf{0.8141}& 0.6833\\
512  & \textbf{0.6133} &  0.8113 & \textbf{0.6837} \\
1024 & 0.6081&0.8008&0.6764 \\
\hline
\end{tabular}
\end{center}
\caption{\label{Batch Size} Impact of the batch size $N$ on the Quora validation set. For computing SDML a batch consists of a paraphrase and $N - 1 $ negative examples.}
\end{table} 

\begin{table}[t!]
\begin{center}
\begin{tabular}{|l|l|rcl|}
\hline \bf Loss & \bf Dist & \bf P@1 & \bf P@10 & \bf MRR \\ \hline
TL (Rand) & EUC & 0.4742 & 0.6509 & 0.5359 \\
TL (Rand) & SSD & 0.5763 & 0.7640 & 0.6421 \\\hline
% K 512 0.1 & SSD & 0.5901 & 0.7841& 0.6591 \\
% K 512 0.1 & EUC & 0.5849  & 0.7733 & 0.6512\\\hline
SDML & SSD &\textbf{0.6133} &  \textbf{0.8113} & \textbf{0.6837}\\
\hline
\end{tabular}
\end{center}
\caption{\label{quora_dev} Comparison of different loss functions on Quora validation set.}
\end{table}

\begin{table}[t!]
\begin{center}
\begin{tabular}{|l|l|rcl|}
\hline \bf Loss & \bf Dist & \bf P@1 & \bf P@10 & \bf MRR \\ \hline
TL (Rand) & EUC & 0.4641 & 0.6523 & 0.5297 \\
TL (Rand) & SSD & 0.5507 & 0.7641 & 0.6265 \\\hline
% K 512 0.1 & SSD & 0.5834 & 0.7939 & 0.6560 \\
% K 512 0.1 & EUC & 0.5746 & 0.7730 & 0.6450 \\\hline
SDML & SSD &\textbf{0.6043} &  \textbf{0.8179} & \textbf{0.6789}\\
\hline
\end{tabular}
\end{center}
\caption{\label{quora_test} Comparison of different loss functions on Quora test set.}
\end{table}

\begin{table}[t!]
\begin{center}
\begin{tabular}{|l|l|rcl|}
\hline \bf Loss & \bf Dist & \bf P@1 & \bf P@10 & \bf MRR \\ \hline
TL (Rand) & EUC &0.5738 &0.7684&0.6428\\
TL (Rand) & SSD  &0.6506 &0.8579 &0.7252\\ \hline
TL (Hard) & EUC &0.5549&0.7534&0.6256\\
TL (Hard) & SSD & 0.5233 & 0.7077 & 0.5870\\ \hline
% K 512 0.1 & SSD & 0.6599 & 0.8777 & 0.7391 \\
% K 512 0.1 & EUC & 0.6471 & \textbf{0.8877 }& 0.7362 \\\hline
SDML & EUC & 0.6526 & \textbf{0.8832} & 0.7361 \\
SDML & SSD & \textbf{0.6745} & 0.8817 & \textbf{0.7491} \\

\hline
\end{tabular}
\end{center}
\caption{\label{od_dev}Comparison of different loss functions on open domain QA dataset validation set.}
\end{table}

\begin{table}[t!]
\begin{center}
\begin{tabular}{|l|l|rcl|}
\hline \bf Loss & \bf Dist & \bf P@1 & \bf P@10 & \bf MRR \\ \hline
TL (Rand) & EUC &0.5721 &0.7695&0.6431\\
TL (Rand) & SSD &0.6538 & 0.8610&0.7271\\ \hline
TL (Hard) & EUC & 0.5593 & 0.7593 & 0.6304\\
TL (Hard) & SSD & 0.5201 & 0.7095 & 0.5863\\ \hline
% K 512 0.1 & SSD & 0.6612 & 0.8780 & 0.7403 \\
% K 512 0.1 & EUC & 0.6490 & \textbf{0.8884} & 0.7368 \\\hline
SDML & EUC & 0.6545 & \textbf{0.8846} & 0.7382 \\
SDML & SSD & \textbf{0.6718} & 0.8830 & \textbf{0.7480} \\

\hline
\end{tabular}
\end{center}
\caption{\label{od_test} Comparison of different loss functions on open domain QA dataset test set.}
\end{table}

Tables ~\ref{od_dev} and \ref{od_test} show the results on the open domain QA dataset validation and test set. TL(Rand) is the triplet loss with random sampling of negative examples whereas TL(Hard) is a variant with hard negative mining. In both the cases the SDML outperforms triplet loss by a considerable margin. It is important to note that since our dataset contains noisy examples triplet loss with random sampling outperforms hard sampling setting, in contrast with the results presented in \citealt{manmatha2017sampling}.

The results presented in this section are consistent with our expectations based on the design of the loss function.

\section{Conclusion}
\label{sec:length}
We investigated a variant of the paraphrase identification task - large scale question paraphrase retrieval, which is of particular importance in industrial question answering applications. We devised a weak supervision algorithm to generate training data from the logs of an existing high precision question-answering system, and introduced a variant of the popular Quora dataset for this task. In order to solve this task efficiently, we developed a neural information retrieval system consisting of a convolutional neural encoder and a fast approximate nearest neighbour search index. 

Triplet loss, a standard baseline for learning-to-rank setting, tends to overfit to noisy examples in training. To deal with this issue we designed a new loss function inspired by label smoothing, which assigns a small constant probability to randomly paired question utterances in a training mini-batch resulting in a model that demonstrates superior performance. We believe that our batch-wise smoothed loss formulation will be applicable to a variety of metric learning and information retrieval problems for which triplet loss is currently popular. The loss function framework we describe is also flexible enough to experiment with different priors - for e.g. allocating probability masses based on the distances between the points.

\bibliography{naaclhlt2019}
\bibliographystyle{acl_natbib}

\end{document}